\documentclass[runningheads]{llncs}
\usepackage[T1]{fontenc}
\usepackage{graphicx}
\usepackage{booktabs}
\usepackage[misc]{ifsym}
\newcommand{\corr}{(\Letter)}
\usepackage{mwe}

\usepackage{comment}

\usepackage{float}
\usepackage{multirow}
\usepackage{algorithm}
\usepackage{algpseudocode}
\usepackage{placeins}
\usepackage[table]{xcolor}
\usepackage{textcomp}
\usepackage{dblfloatfix}
\usepackage{subcaption}

\usepackage{hyperref}
\usepackage{amsmath}
\usepackage{amssymb} 
\usepackage{wrapfig}
\usepackage{subfiles}

\usepackage[color=green!40]{todonotes}

\newcommand{\forecastad}{\texttt{ForecastAD}}
\newcommand{\densityad}{\texttt{DensityAD}}
\newcommand{\xltt}{\texttt{xLTT}}
\newcommand{\drxltt}{\texttt{DR-xLTT}}
\newcommand{\lxltt}{\texttt{L-xLTT}}

\newcommand{\codeurl}{\href{https://github.com/yoest/reliable-ad-csp}{https://github.com/yoest/reliable-ad-csp}}
\newcommand{\simdataurl}{\href{https://tinyurl.com/macmnjyt}{https://tinyurl.com/macmnjyt}}

\begin{document}

\title{Risk-Based Thresholding for Reliable Anomaly Detection in Concentrated Solar Power Plants}
\titlerunning{Risk-Based Thresholding for Reliable Anomaly Detection in CSP Plants}




\author{Yorick Estievenart\inst{1} \and
Sukanya Patra\inst{1} \and
Souhaib Ben Taieb\inst{2} \corr} 

\institute{University of Mons, Mons, Belgium \email{yorick.estievenart@gmail.com,sukanya.patra@umons.ac.be}
\and
Mohamed bin Zayed University of Artificial Intelligence, Abu Dhabi, United Arab Emirates \email{souhaib.bentaieb@mbzuai.ac.ae}}

\tocauthor{Yorick Estievenart, Sukanya Patra, Souhaib Ben Taieb}
\toctitle{Risk-Based Thresholding for Reliable Anomaly Detection in Concentrated Solar Power Plants}



\maketitle              

\begin{abstract}

Efficient and reliable operation of Concentrated Solar Power (CSP) plants is essential for meeting the growing demand for sustainable energy. However, high-temperature solar receivers face severe operational risks, such as freezing, deformation, and corrosion, resulting in costly downtime and maintenance. To monitor CSP plants, cameras mounted on solar receivers record infrared images at irregular intervals ranging from one to five minutes throughout the day. Anomalous images can be detected by thresholding an anomaly score, where the threshold is chosen to optimize metrics such as the F1-score on a validation set. This work proposes a framework, using risk control, for generating more reliable decision thresholds with finite-sample coverage guarantees on any chosen risk function. Our framework also incorporates an abstention mechanism, allowing high-risk predictions to be deferred to domain experts. Second, we propose a density forecasting method to estimate the likelihood of an observed image given a sequence of previously observed images, using this likelihood as its anomaly score. Third, we analyze the deployment results of our framework across multiple training scenarios over several months for two CSP plants. This analysis provides valuable insights to our industry partner for optimizing maintenance operations. Finally, given the confidential nature of our dataset, we provide an extended simulated dataset\footnote{\simdataurl}, leveraging recent advancements in generative modeling to create diverse thermal images that simulate multiple CSP plants. Our code is publicly available\footnote{\codeurl}.


\end{abstract}

\section{Introduction}

The global transition toward greener and more sustainable renewable energy sources is hindered by two critical challenges: (i) on-demand generation and (ii) dispatchability. Concentrated Solar Power (CSP) plants offer a promising solution, leveraging thermal energy storage to provide electricity even when sunlight is unavailable. Among the various CSP configurations, central tower-based plants are the most prevalent, using an array of mirrors to concentrate sunlight onto a receiver, where a heat transfer medium absorbs and stores the energy. However, the extreme operating temperatures make these systems highly susceptible to failures such as metal fatigue and tube blockages, directly impacting their efficiency, reliability, and operational lifespan. To mitigate these risks, thermal imaging from infrared cameras is used to monitor CSP plants. Nonetheless, the sheer volume and complexity of thermal image data render manual monitoring impractical, necessitating the development of an automated, data-driven Predictive Maintenance (PdM) pipeline. This problem naturally aligns with anomaly detection (AD), where the goal is to identify abnormal behaviours.  

Despite significant progress in both deep and shallow AD research \cite{liu2024deep,Ruff2021ADetection}, existing image- and video-based approaches fall short in addressing the problem of detecting anomalous behaviours of operational CSP plants due to three key challenges. First, the lack of interpretability of the anomaly scores hinders decision-making in high-stakes applications without an appropriate thresholding strategy \cite{perini2020quantifying}. Traditional approaches rely on performance metrics such as F1-score or GMean to determine thresholds depending on the available labelled samples. These methods do not guarantee that the results will remain consistent in a deployment setting. Moreover, they assume that all CSP plants define risk similarly and follow the same operational strategies. In reality, this often differs (e.g., deploying a maintenance team may be preferable to replacing a tower component). Second, deep learning-based AD models are often perceived as unreliable \cite{perini2020quantifying} due to the uncertainty in predictions stemming from their inability to properly estimate the decision boundary, particularly when training data is limited. Thus, practitioners are hesitant to use the predictions even when the associated uncertainty is minimal, severely limiting their adoption in real-world applications. Also, unlike classical image- and video-based AD data, CSP plant monitoring involves thermal images without semantic content, lacks a fixed frame rate, and exhibits significant non-stationarity and temporal dependencies due to pronounced daily seasonal patterns. As a result, conventional image- and video-based anomaly detection methods are inappropriate. A recent forecasting-based AD method, \forecastad{} \cite{Patra2024-jk}, attempts to address these challenges by measuring per-pixel errors between predicted and observed thermal images. However, reconstruction-based AD methods suffer a critical flaw: models trained on normal data can inadvertently reconstruct and misclassify anomalous images as normal \cite{Bouman2025-fa,Moore2024-eb}, leading to unreliable detection. 

To overcome these limitations, we propose a principled, robust AD framework tailored for CSP plant monitoring. First, we introduce a risk-controlling thresholding strategy for anomaly scores that satisfies finite-sample performance guarantees on any chosen risk function (e.g., false positive rate or F1-score)—a critical requirement for reliable predictive maintenance (PdM) in industrial settings. To enhance trust and adoption, we integrate a machine-learning-with-abstention framework \cite{Perini2023-zl} with adaptive thresholds that account for the overlap between normal and anomalous score distributions. This approach defers high-risk predictions to domain experts, ensuring human intervention when uncertainty is high. Furthermore, we propose an AD method based on density forecasting, \densityad{}, which leverages conditional normalizing flows to model the likelihood of an observed sample being normal, given past thermal images and timestamps. This approach mitigates the limitations of reconstruction-based methods and enables likelihood-based thresholding for more effective anomaly detection. Our key contributions are: 

\begin{itemize}  
    \item We propose a framework for computing reliable anomaly detection thresholds with finite-sample performance guarantees for any chosen risk function. The framework includes an abstention mechanism that defers decisions to domain experts under high uncertainty.

    \item We develop an unsupervised AD method that computes anomaly scores using density forecasting by estimating the conditional likelihood of an observed infrared image given a sequence of previously observed images.

    \item We conduct an extensive deployment analysis of our framework across multiple real-world scenarios over several months, using data from two CSP plants. This analysis provides valuable insights to our industry partner for maintenance operations.

    \item We release a simulated dataset by leveraging recent advancements in generative modelling to create diverse infrared images that emulate real-world data from CSP plants.  
\end{itemize}

Our work not only advances the state of anomaly detection in renewable energy systems but also serves as an important milestone for future research in robust, data-driven PdM strategies for critical infrastructure monitoring.

\section{Anomaly detection in thermal images from CSP plants}  
\label{sec:usecase}

In the following, we describe our AD use case and the associated dataset.\\

\noindent \textbf{Use-case}. Concentrated Solar Power (CSP) plants are designed to harness solar energy for large-scale electricity generation while addressing two major challenges commonly associated with renewable energy sources -- on-demand generation and dispatchability. Among the four primary CSP technologies currently in use, namely, Solar Tower, Parabolic Trough, Linear Fresnel, and Dish-Stirling systems, this study specifically focuses on the operational aspects of Solar Tower-based CSP plants. These plants comprise two critical components: the Thermal Solar Receiver and the Steam Generator. Positioned atop a central tower, the Solar Receiver functions as a solar furnace, absorbing concentrated sunlight reflected by an array of heliostats—movable mirrors strategically arranged on the ground around the tower. A high-capacity heat transfer medium, such as molten salts, circulates through vertical heat exchanger tubes configured as panels within the receiver, absorbing the thermal energy from the concentrated sunlight. This heated transfer medium is subsequently stored in a Thermal Energy Storage (TES) system. It is later used to generate superheated steam, which drives the Steam Generator to produce electricity. Thus, the incorporation of TES enables the on-demand power generation capability of CSP plants and positions them as viable alternatives to conventional fossil fuel-based power plants.
Despite their advantages, CSP plants encounter significant operational challenges operating in extremely high temperatures. These challenges include blockage or deformation of heat exchanger tubes, metal fatigue, and corrosion, all of which can impact plant efficiency and reliability. Therefore, continuous monitoring and real-time failure detection are crucial to ensuring uninterrupted power generation and preventing costly system failures. In this study, we focus on detecting failures and anomalous behaviours in the Thermal Solar Receiver.\\

\noindent \textbf{Dataset}. As previously discussed, the receiver consists of vertical heat exchanger tubes arranged in panels through which the heat transfer medium flows. During normal operation, the temperature of this medium increases as it moves through the tubes, absorbing heat from concentrated sunlight. It results in a surface temperature gradient along the flow direction, which is captured by infrared (IR) cameras. In this study, our goal is to identify anomalous behaviours of the solar receiver by monitoring these temperature gradients.  
The \emph{solar receiver dataset} used in this study consists of sequences of IR images taken at irregular intervals ranging from one to five minutes throughout the day, with each sequence corresponding to an operational day of the CSP plant. Notably, the dataset lacks ground truth labels, as domain experts do not have prior knowledge of all possible failure types, and anomalies are inherently unknown apriori. Each operational day at the CSP plant comprises three distinct phases: (i) \emph{preheating}, to prevent molten salt from freezing, (ii) \emph{filling/draining}, during which salt circulates at the start and is drained at the end of the operation, and (iii) the \emph{power} phase, where the salt absorbs thermal energy for power generation. Each phase exhibits a distinct surface temperature profile, which must be accounted for in modelling to ensure reliable AD. For example, low surface temperatures are expected during \emph{preheating}, but the same behaviour during the \emph{power} phase may signal a failure. 

\begin{figure}[t]
    \centering
    \includegraphics[width=\linewidth]{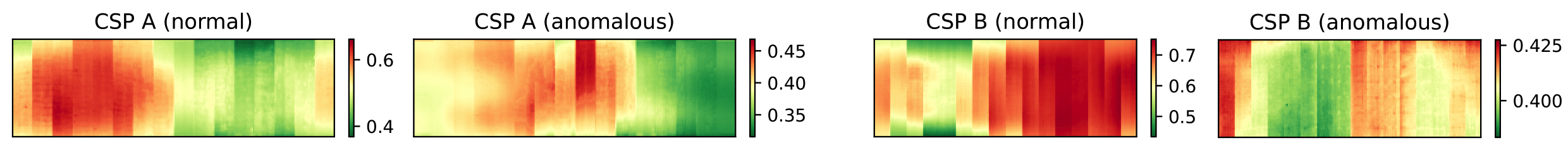}
    \caption{Example of thermal images from CSP $A$ and $B$.}
    \label{fig:samples_csp}
\end{figure}

Building on prior work \cite{Patra2024-jk}, we expanded the \emph{solar receiver dataset} to include data from two distinct CSP plants, referred to as $A$ and $B$, for anonymity. Specifically, we have access to $16343$ samples from CSP $A$ and $15181$ from CSP $B$. Although the dataset exhibits similar key characteristics—such as non-stationarity, irregular sampling, and temporal dependence—certain differences exist across the plants. Notably, the thermal image resolutions differ, with CSP $A$ capturing images of size $184 \times 608$ pixels, while CSP $B$ captures images of size $196 \times 528$ pixels. Furthermore, CSP $B$ exhibits an inversion in the thermal flow direction (left to right), whereas CSP $A$ follow a right-to-left flow pattern. Examples of thermal images for both CSP plants can be seen in Figure \ref{fig:samples_csp}.  To maintain confidentiality, all thermal images have been normalized before analysis.

\section{Background}
\label{sec:general_problem_formulation}

\noindent \textbf{Notations.} We consider an unsupervised AD setting, where the training dataset, denoted as \(\mathcal{D}_{N} = \{x_i\}_{i=1}^n\), consists of \(n\) \textit{unlabeled} samples. Each sample \(x_i = (y_i, t_i) \in \mathcal{X}\) is a tuple, where \(\mathcal{X} = \mathbb{R}^d_+ \times \mathbb{R}_+\). The first component, \(y_i \in \mathcal{Y}\), represents a thermal image of dimension \(d = H \times W\), where \(H\) and \(W\) denote the height and width, respectively, with \(\mathcal{Y} = \mathbb{R}^d_+\). The second component, \(t_i \in \mathbb{R}_+\), corresponds to the timestamp at which the thermal image \(y_i\) was captured. Following prior works \cite{Roth2022TowardsDetection}, we assume that the training dataset \(\mathcal{D}_{N}\) predominantly contains normal samples. Additionally, we introduce another \textit{labelled} dataset, \(\mathcal{D}_{R} = \big\{(x_i, z_i)\big\}_{i=1}^{n_R}\), consisting of \(n_R\) labeled pairs, where \(n_R \ll n\). Each label \(z_i \in \{0,1\}\) indicates whether the corresponding sample is normal (\(z_i = 0\)) or anomalous (\(z_i = 1\)). Furthermore, the dataset \(\mathcal{D}_{R}\) is partitioned into three disjoint subsets: validation (\(\mathcal{D}_V\)), calibration (\(\mathcal{D}_C\)), and test (\(\mathcal{D}_T\)).\\

\begin{figure}[t]
    \centering
    \includegraphics[width=\linewidth]{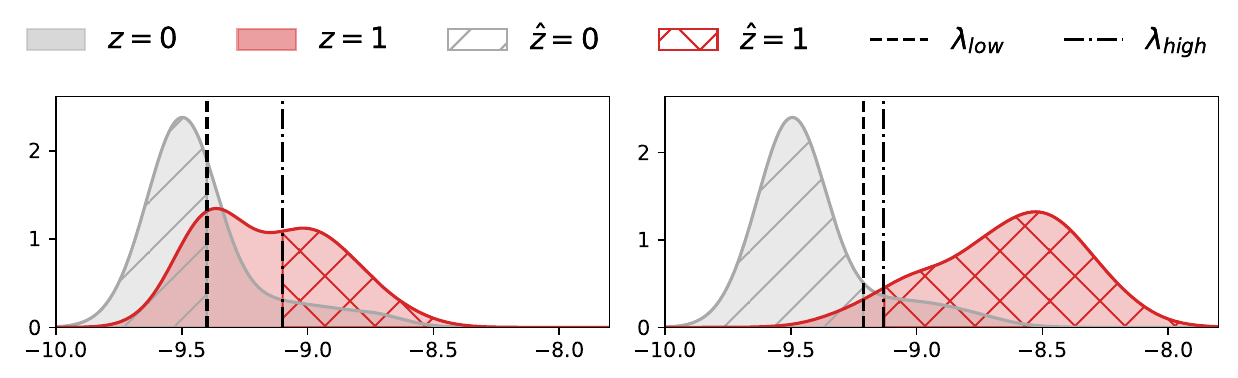}
    \caption{Illustration of thresholds for AD with abstention under high (left) and low (right) overlap in anomaly score distributions of normal and anomalous samples.}
    \label{fig:density_thresholds_example}
\end{figure}

\noindent \textbf{Unsupervised AD.} The goal of unsupervised AD is to estimate an anomaly score function \( s(\cdot): \mathcal{X} \to \mathbb{R} \) using \( \mathcal{D}_N \), such that normal samples receive lower scores. A label (0 for normal or 1 for anomalous) is then assigned to a new test sample \( x \in \mathcal{X} \) by thresholding its anomaly score:
\begin{equation}  
    \label{eq:uad}
    \hat{z} = h(x) = \begin{cases} 
        0, & \text{if } s(x) \leq \lambda, \\ 
        1, & \text{if } s(x) > \lambda, 
    \end{cases}
\end{equation}
where $h : \mathcal{X} \rightarrow \{ 0, 1 \}$ is the labelling function and \( \lambda \in \mathbb{R} \) is a threshold to be determined, whose optimal value depends on the proportion of anomalies in the test set \cite{Perini2022-my,Perini2022TransferringSimilarity}. However, since the true proportion is unknown in practice, existing methods rely on test performance metrics to select a threshold \( \lambda \in \Lambda \) from a set of feasible thresholds \( \Lambda \subset \mathbb{R} \). Commonly adopted approaches include:

\begin{itemize}
    \item \textbf{F1-score \cite{akcay2022anomalib}}. The threshold $\lambda_{\text{F}}$ yields the highest F1-score:
    \begin{equation}
        \lambda_{\text{F}} = \arg \max_{\lambda \in \Lambda} ~  \text{F1-Score}\big( \mathcal{H}_{V}\big),
    \end{equation}
    where $\mathcal{H}_{V} = \{ (h(x), z) \mid (x, z) \in \mathcal{D}_{V} \}$ and F1-Score computes the harmonic mean of precision and recall.

    \item \textbf{G-Mean \cite{Kubat1997-ei}.} The threshold $\lambda_{\text{G}}$ maximizes the G-Mean:
    \begin{equation}
        \lambda_{\text{G}} = \arg \max_{\lambda \in \Lambda} ~  \text{G-Mean}\big( \mathcal{H}_{V} \big),
    \end{equation}
    where $\text{G-Mean}$ computes the geometric mean of precision and recall.
    \item \textbf{Z-score}.  Let \( \mathcal{S}_{V} \) be the set of anomaly scores for normal samples in $\mathcal{D}_{V}$, defined as \( \mathcal{S}_{V} = \{ s(x) \mid (x, 0) \in \mathcal{D}_{V} \} \) with the corresponding mean and standard deviation denoted by \( \mu_{S_{V}} \) and \( \sigma_{S_{V}} \), respectively. The threshold \( \lambda_{\text{z}} \) is set \( k \) standard deviations above \( \mu_{S_{V}} \). Unlike Eq.~\ref{eq:uad}, where the threshold is applied directly to \( s(x) \), here it is applied to the z-scores, defined as \( s_z(x) = \left| \frac{s(x) - \mu_{S_{V}}}{\sigma_{S_{V}}} \right| \).  
\end{itemize}
For a comprehensive discussion on existing methods for selecting \( \lambda \), we refer to \cite{Perini2022-my}. A key limitation of these approaches is that they do not account for uncertainty when the anomaly score distributions of normal and anomalous samples overlap, as illustrated in Figure \ref{fig:density_thresholds_example}. However, given the high-risk nature of AD applications, it is essential to abstain from assigning labels under high uncertainty. This allows domain experts to intervene, reducing the risk of incorrect classifications and ensuring more reliable decision-making. \\

\noindent \textbf{Unsupervised AD with abstention.} To enable abstention from labeling under high uncertainty, we augment the labeling function with an abstention label (\(\text{\textregistered}\)) and introduce two thresholds ($\lambda^l$ and $\lambda^h$), reformulating \(h(x)\) as follows: 
\begin{equation}
    \label{eq:decision_making_process}
    \hat{z} = h(x) = \begin{cases} 
        0, & \text{if } x \in \hat{C}_{\text{nor}}, \quad \hat{C}_{\text{nor}} = \{ x' \in \mathcal{X} \mid s(x') \leq \lambda^l \}, \\ 
        1, & \text{if } x \in \hat{C}_{\text{ano}}, \quad \hat{C}_{\text{ano}} = \{ x' \in \mathcal{X} \mid s(x') \geq \lambda^h \}, \\
        \text{\textregistered}, & \text{if } x \in \hat{C}_{\text{abs}}, \quad \hat{C}_{\text{abs}} = \{ x' \in \mathcal{X} \mid \lambda^l < s(x') < \lambda^h \},
    \end{cases}
\end{equation}
where \( \hat{C}_{\text{nor}}\) is the \textit{normal prediction region}, \( \hat{C}_{\text{ano}} \) is the \textit{anomalous prediction region}, and \( \hat{C}_{\text{abs}}\) is the \textit{abstention region}, where the model refrains from making a decision.

Figure \ref{fig:density_thresholds_example} illustrates two examples of this decision-making process. The parameters \( \lambda^l \) and \( \lambda^h \) define the normal and anomalous prediction regions while also regulating the abstention region, thereby controlling the abstention rate. \\

\noindent Ideally, the pair of thresholds \( (\lambda^l, \lambda^h) \) should adapt to the anomaly score distribution, effectively capturing the overlap between normal and anomalous scores. A trivial yet uninformative approach is to set \( \lambda^l = -\infty \) and \( \lambda^h = +\infty \), which results in abstaining from prediction for all samples. We aim to propose a principled method for selecting a reliable pair of thresholds. 

\section{Reliable Decision Thresholds for AD}
\label{sec:reliable_decision_thresholds}

Let us define a Risk-Controlling Prediction Set (RCPS) \( \widehat{C}_\lambda \) for a given threshold \( \lambda \in \Lambda \subset \mathbb{R} \) as follows:

%
\begin{definition}[RCPS \cite{Bates2021-gj}]
    Let $\lambda \in \Lambda$ be a random variable and $R(\cdot) : 2^{\mathcal{X}} \rightarrow \mathbb{R}$ a risk function. The set $\hat{C}_{\lambda}$ is defined as an $(\alpha, \delta)$-risk-controlling prediction set if it satisfies the condition $\mathbb{P}(R(\hat{C}_{\lambda}) \leq \alpha) \geq 1 - \delta$, where $\alpha \in [0,1]$ is the risk tolerance and $\delta \in [0,1]$ is the error level.
\end{definition}
One method for constructing an RCPS is \emph{conformal risk control}, an extension of conformal prediction (CP) \cite{Angelopoulos2021-st} designed to control the expected value of a risk function, assuming it is monotonically non-increasing with respect to a single threshold \( \lambda \). However, this approach is limited to a single-parameter setting, as in \eqref{eq:uad}, and relies on a restrictive assumption about the risk function. 

To overcome these limitations, we propose leveraging the \emph{Learn then Test} (LTT) procedure \cite{Angelopoulos2021-nz}. We consider the unsupervised AD problem with abstention, as defined in \eqref{eq:decision_making_process}. Our objective is to determine a pair of reliable thresholds \( (\lambda^l, \lambda^h) \) that define a RCPS \( \hat{C}_{(\lambda^l, \lambda^h)} = \hat{C}_{\text{nor}} \cup \hat{C}_{\text{ano}} \) with finite-sample coverage guarantees for any given risk function \( R(\cdot) : 2^{\mathcal{X}} \rightarrow \mathbb{R} \) (e.g., the false positive rate). Additionally, we seek to adapt the abstention rate based on the complexity of the risk function. \\

\noindent \textbf{Our LTT procedure for reliable threshold selection.} We propose an extension of the LTT procedure, denoted as \xltt, which generalizes the framework to consider a pair of thresholds \( (\lambda^l, \lambda^h) \) instead of a single threshold \( \lambda \). The procedure begins by defining a set of paired threshold values, \(\Lambda = \{ (\lambda^l_{(a)}, \lambda^h_{(b)}) \mid a, b \in \{ 1, \dots, m \}, \, \lambda^l_{(a)} \leq \lambda^h_{(b)} \} \). Next, we define the null hypotheses \( \mathcal{H}_j : \hat{R}_{n_C}(\hat{C}_{(\lambda^l_j, \lambda^h_j)}) > \alpha \) for each \( (\lambda^l_j, \lambda^h_j) \in \Lambda \), $j \in \{ 1, \dots, | \Lambda | \}$ and $\alpha \in [0, 1]$, where \( \hat{R}_{n_C}(\cdot) : 2^{\mathcal{X}} \to \mathbb{R} \) is an empirical risk function computed on the calibration set \( \mathcal{D}_C \). Accepting \( \mathcal{H}_j \) indicates that \( (\lambda^l_j, \lambda^h_j) \) does not control the risk. To decide whether to accept or reject \( \mathcal{H}_j \) and thus verify whether the risk is controlled for a given pair \( (\lambda^l_j, \lambda^h_j) \), we compute a valid p-value \( p_j \) for every \( \mathcal{H}_j \) using \( \alpha \). This is achieved via a concentration inequality (e.g., the Hoeffding-Bentkus inequality \cite{Bates2021-gj}). Based on the set of p-values \( P = \{ p_j \}_{j \in \{ 1, \dots, | \Lambda | \} } \), we then select the threshold pairs for which the risk is controlled. Since multiple comparisons increase the likelihood of false positives, a correction function \( \mathcal{A} : P \to P' \) with \( P' \subseteq P \) is required to maintain the desired risk control. For example, we define the set \( \mathcal{O} = \mathcal{A}(P) \subset \Lambda \) using Bonferroni correction as \( \mathcal{A}(P) = \{ (\lambda^l_j, \lambda^h_j) \mid p_j \leq \frac{\delta}{| \Lambda |}, p_j \in P \} \). If \( \mathcal{O} = \emptyset \), we set \( \mathcal{O} = \{ (-\infty, \infty) \} \). Finally, any pair \( (\lambda^l, \lambda^h) \in \mathcal{O} \) ensures that \( \hat{C}_{(\lambda^l, \lambda^h)} \) forms a risk-controlling prediction set. This method enables the use of any risk function in a post-hoc manner (i.e., without requiring retraining of a given anomaly detector), making it particularly valuable for AD in CSP plants with diverse and evolving requirements.\\

\noindent \textbf{Optimal threshold selection for AD.} Now that we have obtained the set \( \mathcal{O} \) of threshold pairs that control the risk, our next objective is to (1) avoid trivial selections where \( \lambda^l = -\infty \) and \( \lambda^h = \infty \), and (2) minimize false positives and false negatives while keeping the abstention rate as low as possible. Let \( \mathcal{I}_1 = \{i \mid z_i = 1\} \) and \( \mathcal{I}_0 = \{i \mid z_i = 0\} \) be the set of indices for anomalous and normal points, respectively. \( \hat{z}_i \) are the predicted labels computed using \eqref{eq:decision_making_process}, with \( i = 1, \dots, |\mathcal{D}_V| \). We propose selecting the optimal thresholds \( \lambda^l_* \) and \( \lambda^h_* \) by computing:
\begin{align*}
&\lambda^l_*, \lambda^h_* \\
&= \arg\min_{\lambda^l, \lambda^h \in \mathcal{O}} 
\underbrace{\frac{|\{i \in \mathcal{I}_1 \mid \hat{z}_i = 0\}|}{|\mathcal{I}_1|}}_{\text{False Negative Rate (FNR)}}
+ \underbrace{\frac{|\{i \in \mathcal{I}_0 \mid \hat{z}_i = 1\}|}{|\mathcal{I}_0|}}_{\text{False Positive Rate (FPR)}}
+ \underbrace{\frac{|\{i \mid \hat{z}_i = \varnothing\}|}{|\mathcal{D}_V|}}_{\text{Abstention Rate}}.
\end{align*}

\noindent \textbf{Density-Based Anomaly Score Functions.} Recent work \cite{novello2025exploring} examined the intrinsic connection between anomaly detection and conformal prediction, demonstrating how insights from each field can mutually enhance the other. Building on this perspective, we leverage recent advancements in CP \cite{Dheur2025-ti} to develop novel anomaly score functions \( s(\cdot) \)\footnote{Hereafter, the score function incorporates contextual information \( c \).} for the labeling function in \eqref{eq:decision_making_process}. These score functions are further integrated with the reliable threshold selection procedure \xltt.

Our framework is based upon an invertible, conditional generative model (e.g., normalizing flows) $\hat{g}: \mathcal{V} \times \mathcal{C} \times \mathbb{R}_+ \to \mathcal{Y}$, where $\mathcal{V}$ is a latent variable with a known distribution and \( \mathcal{C}\) is the space of the conditioning variable. We defer the discussion of the exact model used to Section~\ref{sec:density_based_ad_model}. Formally, $\hat{g}(\hat{g}^{-1}(y ; c, t) ; c, t) = y$ for any $c \in \mathcal{C}$, $y \in \mathcal{Y}$ and \(t \in \mathbb{R}_+\). The invertibility allows us to compute the exact density $\hat{f}(y \mid c, t)$ via the change of variables formula. For a test observation $x =(y, t)$, and given \(\hat{g}\), we consider the following two approaches:
\begin{itemize}
    \item \textbf{\drxltt.} The negative log-likelihood is the score function:
    \begin{equation}
        s_{\text{DR}}(x; c) = - \log( \hat{f}(y \mid c,t)).
    \end{equation}
    \item \textbf{\lxltt.} The second approach is based on an invertible model, following the L-CP method introduced in \cite{Dheur2025-ti}. Unlike the output space \( \mathcal{Y} \), we expect the latent space \( \mathcal{V} \) to be more structured, where normal samples are ideally clustered near the origin. Consequently, in \lxltt{}, we frame the decision-making process as a one-class classification problem in the latent space. Assuming the latent variable follows a standard normal distribution, we use the \( \ell_2 \) distance of the latent representation from the origin as the anomaly score for a test point \( x \):
    \begin{equation}
        s_{\text{L}}(x; c) = \| v \|, \quad \text{where } v = \hat{g}^{-1}(y; c, t).
    \end{equation}
\end{itemize}

\section{Density-based AD Model}
\label{sec:density_based_ad_model}

\begin{figure}[h]
    \centering
    \includegraphics[width=\linewidth]{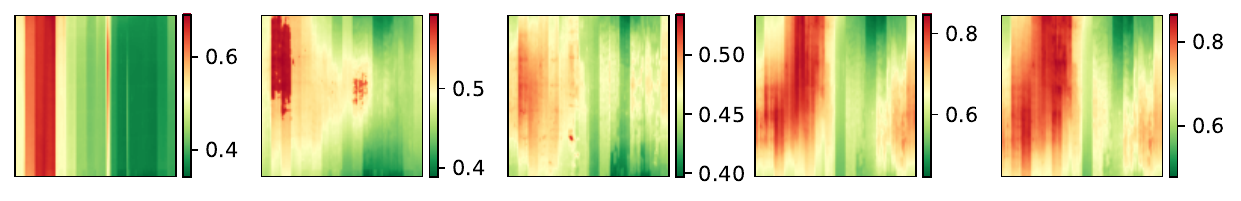}
    \caption{Well-reconstructed anomalous thermal images with empirically low density.}
    \label{fig:anomalies_well_reconstructed}
\end{figure}

The most recent AD model for CSP plants, \forecastad{}, is a reconstruction-based AD methods. However, prior research has shown that anomalies, despite significantly different from normal data, can often be reconstructed in practice \cite{Bouman2025-fa}. For instance, in a bimodal distribution, the distance between the two peaks is greater than the distance between a peak and the local minimum separating them. In such cases, when a prediction aligns with one of the peaks, observations near the local minimum exhibit lower reconstruction errors and thus are incorrectly deemed more likely \cite{Moore2024-eb}. Figure \ref{fig:anomalies_well_reconstructed} presents examples of IR images that are well reconstructed but are anomalous and exhibit empirically low density.

To overcome such limitations of reconstruction-based approaches, we introduce \densityad{}, an invertible generative model that directly estimates the density of thermal images given the contextual information from past images. \densityad{} operates in two main steps: (i) concatenating the $K$ preceding images and their timestamps as a context vector $c$, and (ii) leveraging this context to estimate the density of the current observation $x = (y,t)$, i.e.  $f(y \mid c,t)$.\\

\noindent \textbf{Context encoding.} Building on \cite{Patra2024-jk}, given a test observation \(x_i = (y_i, t_i)\), we construct a rich contextual representation $c_i$ for AD by encoding both spatial and temporal information from the preceding $K$ images. First, at each time step $t_{i-k}$, where \(k =1, \cdots, K\), the corresponding image $y_{i-k}$ is mapped into a lower-dimensional latent space. Specifically, we define an image encoder $\phi_e(\cdot; W_e): \mathcal{Y} \rightarrow \mathcal{V}'$, which transforms images from the high-dimensional input space $\mathcal{Y}$ into a lower dimensional latent space $\mathcal{V}' = \mathbb{R}^{d'}$, where $d' \ll d$. Then, to capture temporal dependencies, we consider two temporal features: the inter-arrival time $\tau_{i-k} = t_{i-k} - t_{i-(k+1)}$, which represents the time elapsed since the previous observation, and the relative time since the start of operation $\gamma_{i-k} = t_{i-k} - t_0$, which situates the observation within the broader operational cycle. These temporal attributes are encoded using a sinusoidal function $\psi(\cdot)$. The final embedding for each data point $(y_{i-k}, t_{i-k})$ is then constructed by concatenating the temporal encodings with the image embedding as $\hat{c}_{i-k} = \phi_e(y_{i-k}; W_e) \oplus \psi_\tau(\tau_{i-k}) \oplus \psi_\gamma(\gamma_{i-k})$. Lastly, to generate the fixed-dimensional context vector $c_i$ at time step $t_i$, the embeddings of the past $K$ images are aggregated using a deep sequence model.\\

\noindent \textbf{Conditional Normalizing Flow.} The conditional PDF \( f(y_i \mid c_i, t_i) \) of the current image \( y_i \), given context \( c_i \) at timestep \( t_i \), is estimated using a conditional normalizing flow, specifically GLOW \cite{Lu2019-ka}. The invertibility property of normalizing flows \cite{rezende2015variational,Kingma2018-ay} enables exact likelihood computation, which is essential for the threshold selection methods discussed in Section~\ref{sec:reliable_decision_thresholds}. To model \( f(y_i \mid c_i, t_i) \), we apply conditional invertible transformations \( g \), mapping \( y_i \) to a latent variable \( v_i \) as \( v_i = g(y_i; c_i, t_i) \). The conditional log-likelihood is then computed using the change-of-variables formula. For further details, we refer the reader to \cite{Lu2019-ka}.

\section{Experiments}  
\label{sec:experiments}

Here, we compare the performance of \densityad{} against existing baselines and assess the efficacy of our proposed decision thresholds for risk-controlled AD. 

\subsection{Experimental Setup}

\noindent \textbf{Dataset.} We use data from two CSP plants, denoted as \( A \) and \( B \). The validation set also serves as a calibration set. For the first data point of each day, both \( \tau \) and \( \gamma \) are initialized to a small positive value, \( \epsilon = 1\text{e-}5 \).

\noindent \textbf{Baselines.} In our evaluation, we compare the performance of \densityad{} against deep image-based AD methods, specifically CFlow \cite{Gudovskiy2021CFLOW-AD:Flows} and DR{\AE}M \cite{Zavrtanik2021DRMDetection}. To extend the comparison to AD approaches that incorporate historical sequences of observations, similar to \densityad{}, we include a spatiotemporal autoencoder (STAE) architecture \cite{hasan2016learning,deepak2021residual,sudhakaran2017learning} and TimeSformer \cite{Bertasius2021-wv}, a transformer-based video classification framework, as baselines, along with \forecastad{} \cite{Patra2024-jk}.

\noindent \textbf{Experimental details.}To prevent numerical instability during training, images are resized to $64 \times 64$, and we employ $3$ flows per block across $5$ blocks. The model is trained using the Adam optimizer with a learning rate of $0.0001$ and a weight decay of $0.00001$. Early stopping is applied based on the validation AUPR\footnote{Area Under the Precision-Recall Curve (AUPR)}, maintaining a fixed balance between normal and anomalous samples in the validation set during training to mitigate the impact of dataset imbalance. The baseline models are trained following their published training setups. We used TimeSformer as the encoder in an encoder-decoder architecture, using the decoder from \forecastad{}, and trained with a mean squared error loss. For the decision thresholds, we use $\alpha = \delta = 0.1$. We conduct an ablation study in Section \(2\) of the supplementary material on the context length \(K\) and the importance of time embeddings \(\tau\) and \(\gamma\). Based on the analysis, we opt for the sequence length $K = 30$ and only consider $\tau$ in \densityad{} for modelling the temporal dynamics.

\noindent \textbf{Evaluation metrics.} We evaluate \densityad{} using two primary metrics: the AUROC\footnote{Area Under the Receiver Operating Characteristic Curve (AUROC)} and the AUPR. Additionally, we assess the proposed thresholding methods by reporting the risk, along with the F1-score and the corresponding abstention rate for two controlled risk measures relevant to our context: the FPR and the F1-score. These choices are not fixed—any risk function can be selected to meet the specific requirements of a CSP plant. We also report these risk measures for existing threshold selection methods. For all experiments, we present the mean over three runs along with one standard error.

\subsection{Results and Discussion}

\noindent \textbf{AD models.} Table \ref{tab:auroc_aupr_results} presents the performance of \densityad{} for both CSP plants. The results indicate that \densityad{} consistently outperforms all baseline methods on both datasets. While STAE, \forecastad{}, and TimeSformer perform well, they still fall short of the performance achieved by our \densityad{}.
\begin{table}[t]
    \caption{AUROC and AUPR performances of \densityad{} against baseline methods. Style: best in \textbf{bold}, and second best \underline{underlined}.}
    \label{tab:auroc_aupr_results}
    \centering
    \aboverulesep = 0pt
    \belowrulesep = 0pt
    \begin{tabular}{@{}c|l|ccc|ccc@{}}
    \toprule
        \textbf{CSP} & \textbf{Model} & \textbf{AUROC (\%)} $\uparrow$ & \textbf{AUPR (\%)} $\uparrow$ \\
        \midrule
    \multirow{6}{*}{A} & CFlow \cite{Gudovskiy2021CFLOW-AD:Flows} & 76.46 $\pm$ 0.92 & 70.32 $\pm$ 1.20 \\
     & DR{\AE}M \cite{Zavrtanik2021DRMDetection} & 81.55 $\pm$ 1.9 & 74.8 $\pm$ 2.79 \\
     & STAE & \underline{89.47 $\pm$ 1.59} & 87.38 $\pm$ 2.4 \\
     & TimeSformer \cite{Bertasius2021-wv} & 87.8 $\pm$ 2.46 & 83.36 $\pm$ 3.15 \\
     & \forecastad{} \cite{Patra2024-jk} & 86.28 $\pm$ 1.74 & \underline{87.57 $\pm$ 1.38} \\
     \rowcolor{blue!6} \cellcolor{white} & \densityad & \textbf{94.25 $\pm$ 0.2} & \textbf{93.88 $\pm$ 0.48} \\
    \midrule
    \multirow{6}{*}{B} & CFlow \cite{Gudovskiy2021CFLOW-AD:Flows} & 55.8 $\pm$ 5.47 & 57.56 $\pm$ 4.85 \\
     & DR{\AE}M \cite{Zavrtanik2021DRMDetection} & 78.82 $\pm$ 5.72 & 71.75 $\pm$ 8.56 \\
     & STAE & \underline{89.9 $\pm$ 1.18} & 88.98 $\pm$ 1.68 \\
     & TimeSformer \cite{Bertasius2021-wv} & 88.59 $\pm$ 2.14 & \underline{89.84 $\pm$ 1.29} \\
     & \forecastad{} \cite{Patra2024-jk} & 81.76 $\pm$ 0.7 & 82.88 $\pm$ 1.39 \\
    \rowcolor{blue!6} \cellcolor{white} & \densityad & \textbf{91.93 $\pm$ 0.52} & \textbf{90.66 $\pm$ 0.46} \\
        \bottomrule
    \end{tabular}
\end{table}

\begin{figure}[t]
    \centering
    \includegraphics[width=\linewidth]{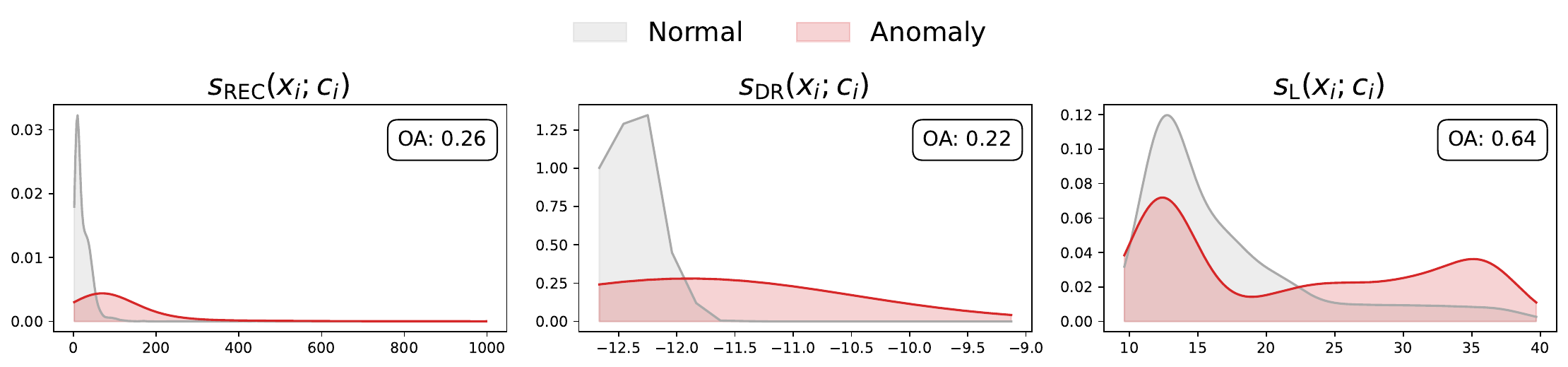}
    \caption{Empirical score distributions of normal and anomalous test samples from CSP $A$ for our proposed score functions and the one used by \forecastad{}, with the overlapping area (OA) between both distributions in the top right corner.}
    \label{fig:scores_insights}
\end{figure}

\noindent \textbf{Anomaly scores.} Figure \ref{fig:scores_insights} shows the distributions of normal and anomalous scores for test samples on CSP $A$, using the proposed scores, introduced in Section~\ref{sec:reliable_decision_thresholds} (i.e., $s_{\text{DR}}$ and $s_{\text{L}}$) and the reconstruction score $s_{\text{REC}}$ from \forecastad{}. In this example, $s_{\text{DR}}$ and $s_{\text{REC}}$ scores effectively distinguish normal from anomalous samples, as shown by the overlapping area (OA) between both distributions.

\begin{figure}[t]
    \centering
    \begin{subfigure}{\textwidth}
        \centering
        \includegraphics[width=\linewidth]{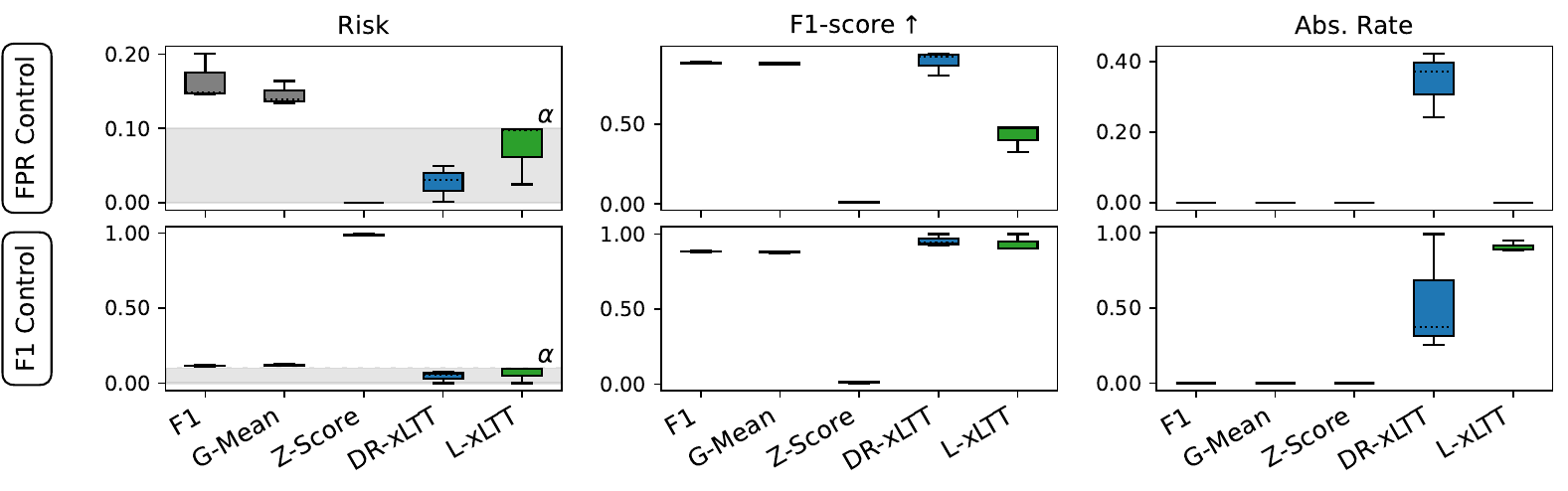}
        \caption{For CSP $A$}
    \end{subfigure}
    \begin{subfigure}{\textwidth}
        \centering
        \includegraphics[width=\linewidth]{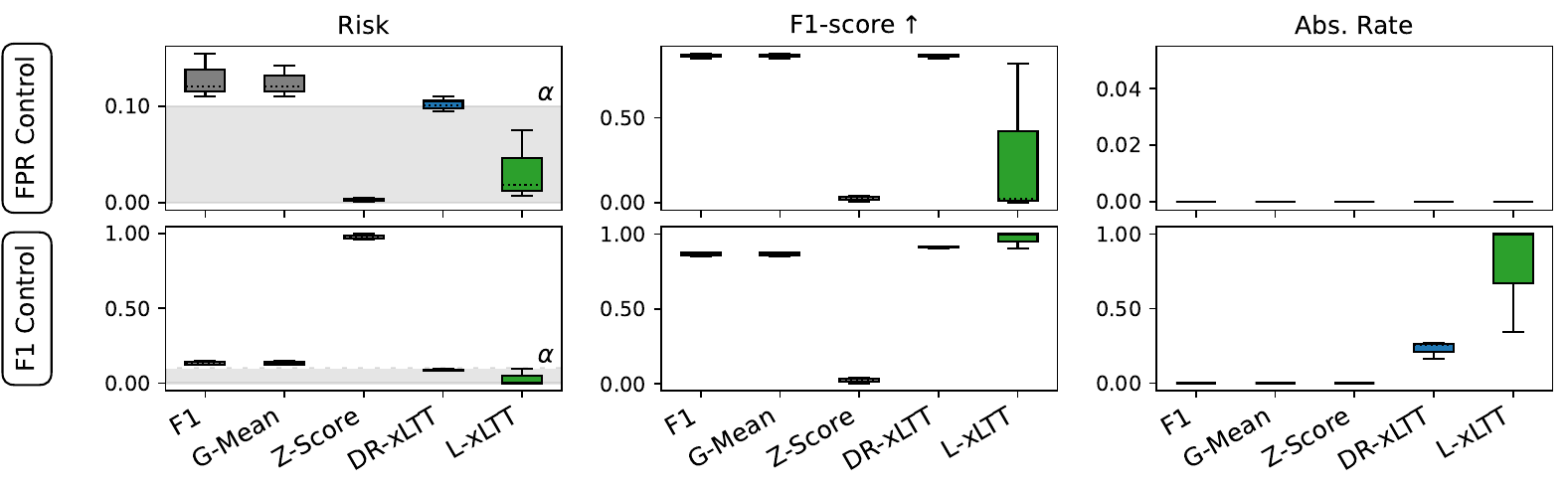}
        \caption{For CSP $B$}
    \end{subfigure}
    \caption{Risk control over FPR (top row) and F1-score (bottom row) for existing and proposed methods. The risk is FPR (top row) and $1 - $F1 (bottom row).}
    \label{fig:thresholding_results}
\end{figure}

\noindent \textbf{Anomaly threshold selection.} Figure \ref{fig:thresholding_results} provides an overview of the threshold selection approaches. The results clearly show that the proposed methods effectively control risk for both risk functions, whereas existing methods do not offer such guarantees. The \drxltt{} methods demonstrate strong performance, balancing risk control with a high F1-score while maintaining a low abstention rate. Notably, they outperform approaches that select the maximum validation set value. Furthermore, these methods adapt to the complexity of the risk function, recognizing that controlling the F1-score presents greater predictive challenges than the FPR. They also fully adjust to user requirements, increasing the abstention rate when constraints are too stringent (e.g., attempting to control the F1-score with a weak underlying model).\\

\noindent \textbf{Computation requirements.} The models are trained using a single NVIDIA A100 GPU with 40 GB of memory, along with 8 CPU cores and 20 GB of RAM. Table \ref{tab:compute_time} presents the training times (excluding the risk-control), memory usage and inference time for a single test point. As shown in Table \ref{tab:auroc_aupr_results}, \densityad{} performs better than \forecastad{} while using similar resources.

\begin{table}[t]
    \caption{Total CPU time and memory used for training the models.}
    \label{tab:compute_time}
    \setlength{\tabcolsep}{6pt} 
    \aboverulesep = 0pt
    \belowrulesep = 0pt
    \centering
    \begin{tabular}{c|cc|cc}
        \toprule
        & \multicolumn{2}{c|}{\forecastad{}} & \multicolumn{2}{c}{\densityad{}} \\
        & CSP A & CSP B & CSP A & CSP B \\
        \midrule
        Training time (s) & 4151 $\pm$ 327 & 2204 $\pm$ 334 & 3760 $\pm$ 836 & 3248 $\pm$ 411 \\
        Memory used (Gb) & 1.63 $\pm$ 0.13 & 1.73 $\pm$ 0.05 & 1.73 $\pm$ 0.02  & 1.63 $\pm$ 0.02 \\
        Inference time (ms) & 194 $\pm$ 6.73 & 177 $\pm$ 1.44 & 201.3 $\pm$ 17.41 & 178 $\pm$ 2.37 \\
        \bottomrule
    \end{tabular}
\end{table}

\begin{figure}[t]
    \centering
    \includegraphics[width=\linewidth]{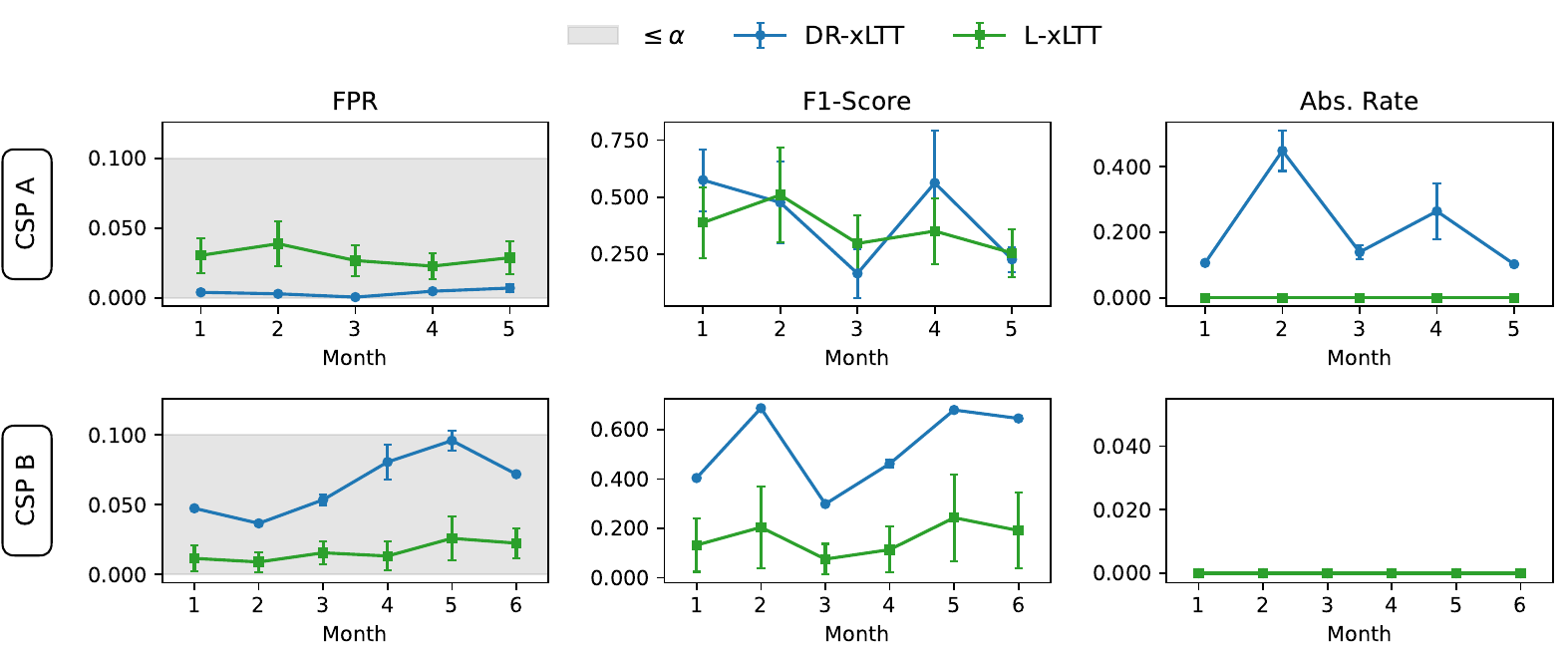}
    \caption{FPR control for the proposed approaches in a deployment setting over multiple months, for the two CSP plants.}
    \label{fig:deployment_thresholding_results}
\end{figure}

\begin{figure}[t]
    \centering
    \includegraphics[width=\linewidth]{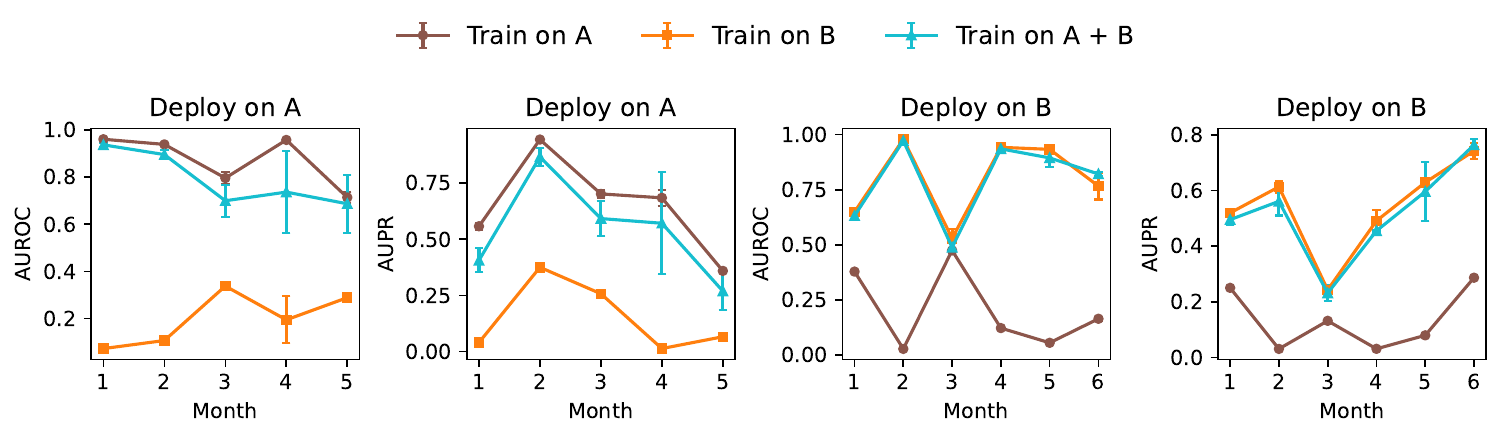}
    \caption{AUROC and AUPR for the two CSP plants over multiple months using three different training settings (i.e. training on $A$, $B$, and $A + B$).}
    \label{fig:deployment_auroc_aupr_results}
\end{figure}

\section{Deployment}  
\label{sec:deployment}

We deployed our threshold selection methods using \densityad{} on 5 and 6 months of anonymized data from CSP plants A and B, respectively. Figure \ref{fig:deployment_thresholding_results} presents the thresholding results, where the FPR is used as the controlled risk. The results demonstrate that risk is effectively controlled in deployment, with \drxltt{} emerging as the most consistent method across both CSP plants. All methods maintain a low abstention rate, making them well-suited for deployment. Additionally, the deployment results align closely with those observed during testing. Performance fluctuations across months can be attributed to variations in the frequency and complexity of anomalies, with some months exhibiting a higher occurrence or more challenging cases.

Figure~\ref{fig:deployment_auroc_aupr_results} evaluates the performance of \densityad{} in deployment under three training configurations: training on CSP A, training on CSP B, and training on a combination of data from both CSP plants. As expected, deploying a model trained on a different tower results in a performance decline. Furthermore, training on data from both plants does not yield any performance improvement, suggesting that information from one tower does not generalize well to another. Although thermal flow patterns are similar across CSPs, anomaly definitions vary due to site-specific factors such as geographic location and operational context. This limits cross-site generalization, indicating the need for per-site models or fine-tuning. Future work could address this through domain adaptation techniques.
Although originally not designed for this purpose, \densityad{} offers a general framework that can be extended to multivariate time series anomaly detection. In this work, we focus on its application to anomaly detection in CSPs, where the anomaly score is computed and subsequently processed through a thresholding mechanism.
Finally, the results suggest that the proposed method supports practical deployment by enabling control over operational risk. For instance, it allows organizations to meet predefined detection targets—such as identifying 90\% of anomalies—thereby supporting compliance with regulatory or performance requirements.

\begin{figure}[t]
    \centering
    \includegraphics[width=\linewidth]{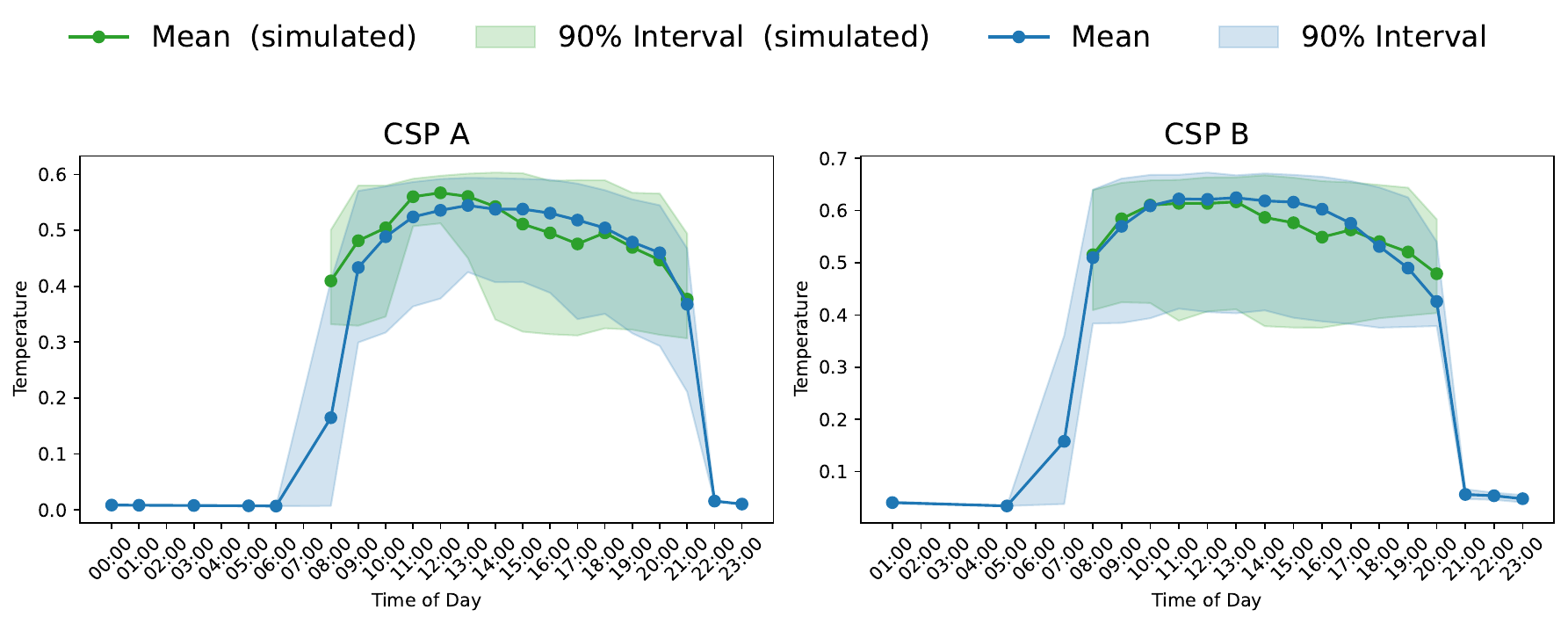}
    \caption{Mean daily temperature for the original and simulated datasets. The shaded area represents the 90\% temperature interval.}
    \label{fig:simulated_data_day_temperature}
\end{figure}

\section{Simulated dataset}  
\label{sec:simulated_dataset}

Building on the methodology described in \cite{Patra2024-jk}, we construct a simulated dataset to facilitate the reproducibility and validation of our results. \densityad{} enables exact likelihood computation while also allowing sampling from the learned distribution. Leveraging this capability, we generate high-quality samples using our proposed density-based model. The dataset simulates two distinct CSP setups, providing a valuable resource for advancing anomaly detection research in CSP plants. Further details are provided in the supplement. 

Due to transformations applied during anonymization, we assessed the reliability of the generated images by comparing the average daily temperature profiles of both CSP plants. As shown in Figure \ref{fig:simulated_data_day_temperature}, temperature levels during the critical period (08:00 to 20:00) closely match between the simulated and original datasets. Temperatures outside this interval, which are notably lower, were regarded as trivial outliers and excluded from the simulated data.

\section{Related work}  
\label{sec:related_work}

\noindent \textbf{Unsupervised AD.} Based on the assumption of a ``clean'' training dataset, i.e., containing only normal samples, unsupervised AD approaches have been proposed with the aim of training models that learn a ``compact'' representation of the normal behaviour. Then, anomalies are identified as deviations from this learned normality. Existing methods can be broadly categorized into four families \cite{Ruff2021ADetection}. First, both deep and shallow \emph{one-class classifiers} \cite{Tax1999SupportDescription,Ruff2018DeepClassification} learn a decision boundary around normal samples with classical methods such as support vector data description. Second, \emph{feature embedding-based} methods store or learn normal data representations using pre-trained models \cite{Roth2022TowardsDetection,Lee2022CFA:Localization} or student-teacher networks \cite{Zhang2023ContextualDetection,Batzner2024EfficientAD:Latencies,patra2024revisiting}. Third, \emph{reconstruction-based} methods use encoder-decoder architectures to map normal samples into a lower-dimensional bottleneck and reconstruct them with high fidelity. Lastly, \emph{density-based} methods estimate the probability distribution of normal samples under the \emph{concentration assumption}, where anomalies are expected to be in low-density regions. For a comprehensive survey, we refer readers to \cite{liu2024deep,Ruff2021ADetection}.


Beyond image-based AD, prior research also investigated AD in videos (VAD) \cite{chandrakala2023anomaly,le2023attention,abdalla2024video}, using historical sequences of observations to identify deviations. However, our setting differs in two key ways: (1) our IR images lack the semantic content of typical video frames, and (2) our solar dataset is captured at irregular intervals, while videos are captured at a fixed frame rate. Although \cite{Patra2024-jk} introduced a forecast-based AD approach for CSP plants, it lacks a reliable selection of AD threshold. Moreover, their study is limited to a single CSP plant, whereas our setting involves multiple plants, introducing additional heterogeinity.\\

\noindent \textbf{Anomaly detection thresholds.} To assign labels, AD methods typically threshold anomaly scores. Commonly used decision thresholds particularly relevant to our use case involve optimizing performance metrics, such as the F1-score \cite{akcay2022anomalib}, G-Mean \cite{Kubat1997-ei}, or the area under the Precision-Recall Curve (PRC), on the validation set over a range of possible thresholds. Another class of methods builds on conformal prediction (CP) \cite{Vovk2005-ur}, a distribution-free framework for constructing prediction sets (e.g., those defined in our decision-making process) providing a finite-sample coverage guarantee. \cite{Bates2021-cp} introduces a method for computing conditionally valid conformal p-values for nonparametric outlier detection, framing the problem within a multiple hypothesis testing context. A key extension of CP, named \emph{conformal risk control} \cite{Angelopoulos2022-am}, shifts the guarantee from coverage to managing any monotonically non-increasing risk function. The \emph{Learn then Test} procedure \cite{Angelopoulos2021-nz} further allows us to extend this concept to any risk function, irrespective of its monotonicity, to generate risk-controlled prediction sets \cite{Bates2021-gj}.

\section{Conclusion}  
\label{sec:conclusion}

We introduced a principled and robust framework for anomaly detection (AD) designed to monitor CSP plants using infrared imagery captured at irregular intervals throughout the day. Our approach labels images as normal or anomalous by first assigning an anomaly score using a model trained on an unlabeled image dataset, followed by a thresholding procedure. To address the challenges of unsupervised AD for CSP plants, our contributions are fourfold. First, we proposed a framework for computing reliable anomaly detection thresholds with finite-sample risk coverage guarantees for any chosen risk function while allowing deferral to domain experts under high uncertainty. Second, to compute more robust anomaly scores for an observed image, we developed a density forecasting method that estimates its likelihood conditional on a sequence of previously observed images. Third, we conducted an extensive real-world deployment analysis over several months across two operational CSP plants, providing valuable insights for industrial maintenance. Lastly, we released a simulated dataset leveraging recent advancements in generative modeling, facilitating data-driven predictive maintenance (PdM) for critical infrastructure. By enhancing the reliability of renewable energy systems, our work supports the broader adoption of sustainable energy solutions for a greener future.

\begin{credits}
\subsubsection{\ackname}
This research is funded by the project ``Federated Learning and Augmented Reality for Advanced Control Centers.'' Special thanks to Thibault GEORGES and Adrien FARINELLE from John Cockerill for their assistance in analyzing the dataset and identifying related abnormal behaviors. We also acknowledge Victor DHEUR for his valuable feedback on the risk control approach.

\end{credits}





\author{Yorick Estievenart\inst{1} \and
Sukanya Patra\inst{1} \and
Souhaib Ben Taieb\inst{2} \corr} 

\institute{University of Mons, Mons, Belgium \email{yorick.estievenart@gmail.com,sukanya.patra@umons.ac.be}
\and
Mohamed bin Zayed University of Artificial Intelligence, Abu Dhabi, United Arab Emirates \email{souhaib.bentaieb@mbzuai.ac.ae}}







\title{Supplementary Material for "Risk-Based Thresholding for Reliable Anomaly Detection in Concentrated Solar Power Plants"}
\titlerunning{Risk-Based Thresholding for Reliable Anomaly Detection in CSP Plants}

\maketitle

\section{Confidentiality Statement}

Our solution has been deployed at two CSP plants, with results over multiple periods presented in this paper to demonstrate its robustness. Due to the proprietary nature of the original data, we cannot make it publicly available. Instead, we have generated a simulated dataset that closely mirrors the characteristics of the CSP plants using advanced data generation techniques, ensuring fidelity while preserving confidentiality. Additionally, a portion of our codebase is now available in an open repository, allowing the research community to experiment with the simulated dataset.

\section{Ablation Study}  \label{appendix:ablation_study}

\noindent \textbf{Impact of image reduction.} In our methodology, we resized the images $256 \times 256$ pixels to $64 \times 64$ to stabilize training. To assess information loss, we evaluate it on a subset of $1000$ samples from our dataset. Specifically, we compute the Mean Squared Error (MSE) relative to the original $256 \times 256$ images, along with the Structural Similarity Index (SSIM) and Peak Signal-to-Noise Ratio (PSNR). Table \ref{tab:abla_study_image_reduction} shows that despite the MSE increasing significantly at lower resolutions, the SSIM remains strong, implying that even low-resolution images retain recognizable structural features. We select $64 \times 64$ as an optimal balance between computational efficiency and image quality.
\begin{table}[h!]
    \caption{Loss of information after reducing the original image ($256 \times 256$).}
    \label{tab:abla_study_image_reduction}
    \aboverulesep = 0pt
    \belowrulesep = 0pt
    \centering
    \begin{tabular}{c|ccc}
        \toprule
        Image Size & MSE & SSIM & PSNR \\
        \midrule
        $128 \times 128$ & $26.0474$ & $1.0000$ & $76.9805$ \\
        \rowcolor{blue!6} $64 \times 64$ & $86.8460$ & $0.9999$ & $71.8129$ \\
        $32 \times 32$ & $209.0905$ & $0.9998$ & $68.0073$ \\
        $16 \times 16$ & $619.8050$ & $0.9994$ & $63.3100$ \\
        \bottomrule
    \end{tabular}
\end{table}

\noindent \textbf{Importance of sequence length.}  The performance of \densityad{} for different sequence lengths \( K \) is shown in Table \ref{tab:abla_seq}. All tested values of \( K \) yield strong results. We choose \( K = 30 \) as the baseline for our experiments, as it provides an optimal balance between capturing sufficient information and avoiding unnecessary complexity.

\begin{table*}[htb]
    \caption{Ablation of $K$. Style: best in \textbf{bold}.}
    \label{tab:abla_seq}
    \centering
    \aboverulesep = 0pt 
    \belowrulesep = 0pt
    \begin{tabular}{@{}c|c|ccc|ccc@{}}
        \toprule
        CSP & $K$ & AUROC (\%) & AUPR (\%) \\
        \midrule
    \multirow{5}{*}{A} & 1 & 92.46 $\pm$ 1.01 & 91.41 $\pm$ 1.42 \\
    & 10 & 93.99 $\pm$ 0.36 & 93.54 $\pm$ 0.51 \\
    & 20 & 93.62 $\pm$ 0.16 & 93.2 $\pm$ 0.24 \\
    \rowcolor{blue!6} \cellcolor{white} & 30 & 92.89 $\pm$ 0.41 & 91.72 $\pm$ 0.65 \\
    & 40 & \textbf{94.62 $\pm$ 0.48} & \textbf{94.3 $\pm$ 0.6} \\
    \cline{1-4}
    \multirow{5}{*}{B} & 1 & 91.43 $\pm$ 0.31 & 90.41 $\pm$ 0.34 \\
    & 10 & 91.08 $\pm$ 0.28 & 90.52 $\pm$ 0.17 \\
    & 20 & 91.3 $\pm$ 0.28 & 90.39 $\pm$ 0.62 \\
    \rowcolor{blue!6} \cellcolor{white} & 30 & \textbf{91.49 $\pm$ 0.1} & \textbf{90.52 $\pm$ 0.24} \\
    & 40 & 90.36 $\pm$ 0.56 & 89.47 $\pm$ 1.05 \\
    \bottomrule
    \end{tabular}
\end{table*}

\noindent \textbf{Importance of time-embedding.} Table \ref{tab:abla_time_emb} provides insights into the necessity of using both $\tau$ and $\gamma$ during training, or whether using only one of them would suffice. It is evident that utilizing only $\tau$ improves performance, and therefore we selected this approach.

\begin{table*}[htb]
    \caption{Ablation of time-embeddings. Style: best in \textbf{bold}.}
    \label{tab:abla_time_emb}
    \centering
    \aboverulesep = 0pt 
    \belowrulesep = 0pt
    \begin{tabular}{@{}c|cc|ccc|ccc@{}}
        \toprule
        CSP & $\tau$ & $\gamma$ & AUROC (\%) & AUPR (\%) \\
        \midrule
    \rowcolor{blue!6} \multirow{3}{*}{A} \cellcolor{white}  & $\checkmark$ & - & \textbf{94.25 $\pm$ 0.2} & \textbf{93.88 $\pm$ 0.48} \\
     & - & $\checkmark$ & 93.05 $\pm$ 0.58 & 92.44 $\pm$ 0.64 \\
    & $\checkmark$ & $\checkmark$ & 94.0 $\pm$ 0.15 & 93.69 $\pm$ 0.22 \\
    \midrule
    \rowcolor{blue!6} \multirow{3}{*}{B} \cellcolor{white}  & $\checkmark$ & - & \textbf{91.93 $\pm$ 0.52} & \textbf{90.66 $\pm$ 0.46} \\
     & - & $\checkmark$ & 90.54 $\pm$ 0.63 & 90.05 $\pm$ 0.72 \\
     & $\checkmark$ & $\checkmark$ & 90.95 $\pm$ 0.13 & 90.58 $\pm$ 0.23 \\
        \bottomrule
    \end{tabular}
\end{table*}

\begin{figure}[h]
    \centering
    \includegraphics[width=\linewidth]{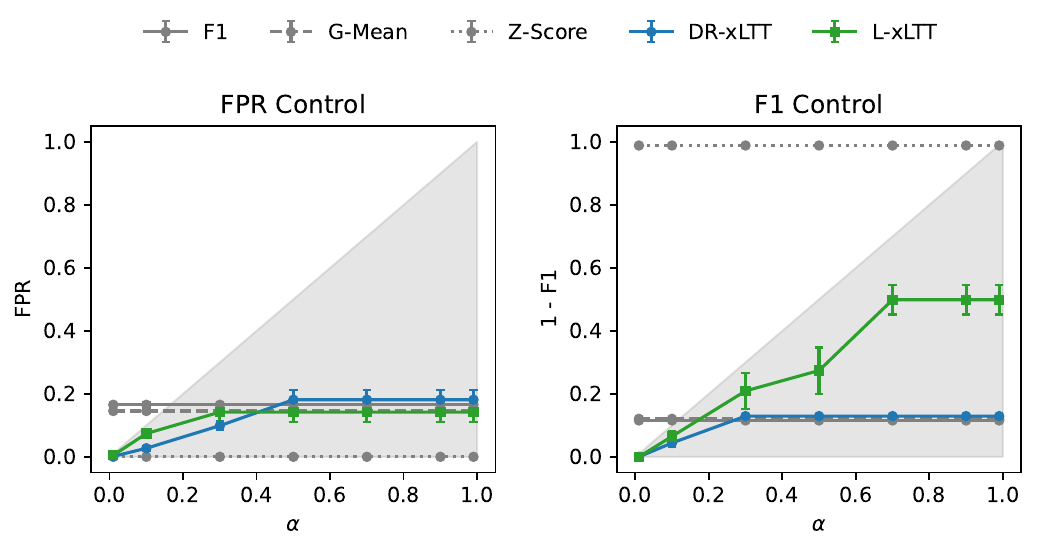}
    \caption{Controlled risks (i.e., FPR and $1 - $F1) across multiple $\alpha$ values, with $\delta = 0.1$.}
    \label{fig:alpha_abla}
\end{figure}

\noindent \textbf{Effect of $\alpha$ on controlled risk.} Figure \ref{fig:alpha_abla} shows the FPR and F1 control for various $\alpha$ values across the proposed threshold selection methods, with $\delta = 0.1$. The results clearly demonstrate that the proposed methods effectively control risk regardless of the $\alpha$ value, whereas the existing methods do not.

\section{Details on dataset simulation}

In this study, having access to multiple CSP plants allowed us to extend the simulated dataset proposed by \cite{Patra2024-jk} to include thermal images from various CSP facilities. This section presents details on this simulated dataset. Each folder corresponds to a CSP plant ($A$ or $B$), with individual samples stored as pickle files named after their respective timestamps. Each file contains a thermal image, its label, and the associated setting (i.e., \emph{Starting} (S), where the mean temperature of the solar receiver begins to rise, \emph{Middle} (M), where it reaches and maintains its peak, or \emph{Ending} (E), where it declines as the day concludes.).

\subsection{Description and performances.} 

\begin{figure}[h]
    \centering
    \includegraphics[width=0.8\linewidth]{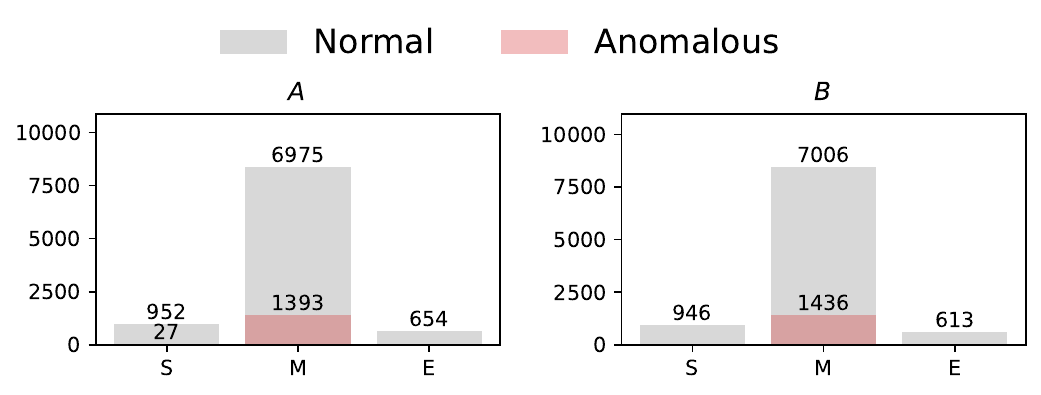}
    \caption{Spread of normal and anomalous samples across each phases for the simulated data of each CSP plant.}
    \label{fig:simulated_data_csp_distribution}
\end{figure}

\begin{figure}[h]
    \centering
    \begin{minipage}[t]{\textwidth}
        \centering
        \includegraphics[width=0.49\linewidth]{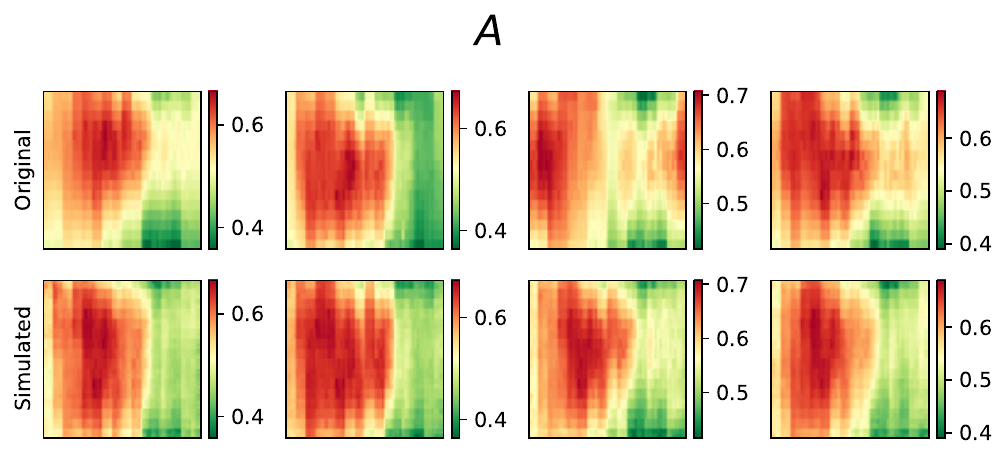}
        \includegraphics[width=0.49\linewidth]{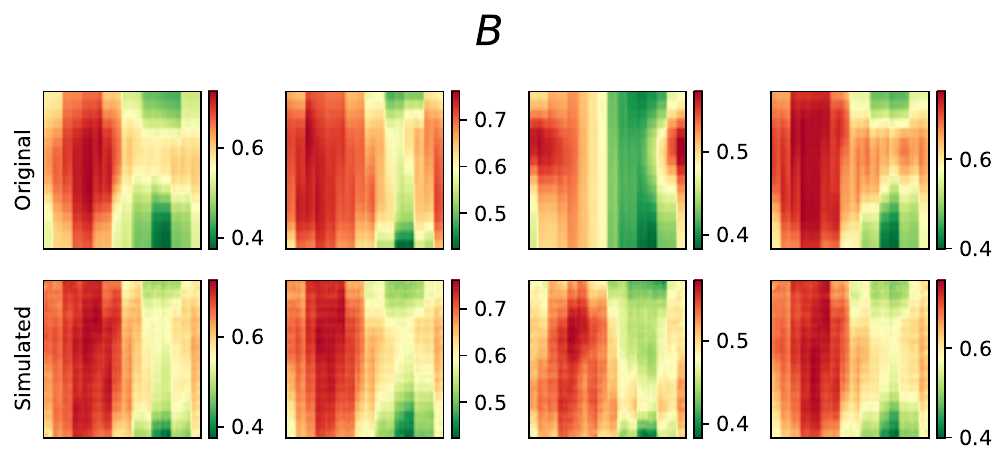}
        \caption{Example of simulated thermal images for the two CSP plants (i.e. $A$ and $B$), along with original image from which it has been sampled.}
        \label{fig:simulated_thermal_images_example}
    \end{minipage}
\end{figure}

The resulting dataset contains $10001$ samples for CSP $A$ and $10001$ samples for CSP $B$, totaling $20002$ samples. The distribution of samples across the \emph{Starting}, \emph{Middle}, and \emph{Ending} phases is illustrated in Figure \ref{fig:simulated_data_csp_distribution}, while examples of simulated samples for each CSP plant are shown in Figure \ref{fig:simulated_thermal_images_example}. The samples closely resemble the real dataset for both CSP plants. However, for anonymity purposes, the data has been resized to $64 \times 64$, normalized between 0 and 1, and anonymized timestamps are used. Additionally, the generation model may still produce some blurry samples. These limitations should be considered when modeling.

The results of applying \densityad{} to this simulated dataset are presented in Figure \ref{tab:auroc_aupr_simulated_dataset_results}. The dataset was divided into training, validation, and test samples. Due to the limited number of samples, we reduced the number of blocks to 3 to mitigate training instability. The method achieves strong performance on CSP \( B \), while performance drops on CSP \( A \). This decline can be attributed to the reduced number of blocks and the greater diversity of samples in CSP \( A \).

\begin{table}[H]
    \caption{AUROC and AUPR of \densityad{} on the simulated dataset.}
    \label{tab:auroc_aupr_simulated_dataset_results}
    \aboverulesep = 0pt
    \belowrulesep = 0pt
    \centering
    \begin{tabular}{cc|cc}
        \toprule
        \multicolumn{2}{c|}{$A$} & \multicolumn{2}{c}{$B$} \\
        AUROC (\%) & AUPR (\%) & AUROC (\%) & AUPR (\%) \\
        \midrule
        75.13 $\pm$ 1.23 & 74.45 $\pm$ 0.66 & 88.36 $\pm$ 0.14 & 91.10 $\pm$ 0.09 \\
        \bottomrule
    \end{tabular}
\end{table}
\FloatBarrier


\end{document}